\documentclass[12pt,a4paper,reqno]{article}

\newcommand{\red}{\ensuremath{\mathcal{P}}}
\newcommand{\rec}{\ensuremath{\mathcal{R}}}

\usepackage{amsmath}
\usepackage{amsfonts}
\usepackage{amssymb}
\usepackage{amsthm}
\usepackage{dsfont}

\usepackage{algorithm}
\usepackage{algpseudocode}

\usepackage{geometry}
\usepackage{graphicx}
\usepackage[unicode,bookmarks,colorlinks=true,breaklinks=true]{hyperref}
\usepackage[utf8]{inputenc}
\usepackage{MnSymbol}
\usepackage[numbers,square]{natbib}
\usepackage{stmaryrd}
\usepackage{subcaption}
\usepackage{tikz}
\usepackage[mode=buildnew]{standalone}

\usepackage{nicematrix}
\usepackage{pgfplots}

\usepackage{smartdiagram}
\usesmartdiagramlibrary{additions}
\usetikzlibrary{shapes.geometric}

\usepackage{comment}

\usepackage{mathtools}
\mathtoolsset{showonlyrefs}

\newtheorem{theorem}{Theorem}
\newtheorem{proposition}{Proposition}
\newtheorem{definition}{Definition}

\definecolor{morange}{RGB}{255,127,14}
\definecolor{mblue}{RGB}{31,119,180}
\definecolor{mred}{RGB}{214,39,40}
\definecolor{mpurple}{RGB}{148,103,189}
\definecolor{mgreen}{RGB}{44,160,44}

\newcommand*{\colIn}{morange}%
\newcommand*{\colEnc}{mpurple}%
\newcommand*{\colIntr}{mred}%
\newcommand*{\colDec}{mpurple}%
\newcommand*{\colOut}{mgreen}%
\newcommand*{\colSym}{mblue}%

\newcommand*{\arrowLength}{1}%

\newcommand*{\xMin}{1}%
\newcommand*{\xMax}{8}%
\newcommand*{\xMaxEnc}{4}%
\newcommand*{\xMaxIntr}{2}%
\newcommand*{\xMaxDec}{4}%

\newcommand*{\xin}{0}%
\newcommand*{\xinsym}{\xin+\arrowLength+2}%
\newcommand*{\xenc}{\xinsym+\arrowLength+2}%
\newcommand*{\xencsym}{\xenc+\arrowLength+2}%
\newcommand*{\xintr}{\xencsym+\arrowLength+2}%
\newcommand*{\xdecsym}{\xintr+\arrowLength+2}%
\newcommand*{\xdec}{\xdecsym+\arrowLength+2}%
\newcommand*{\xoutsym}{\xdec+\arrowLength+2}%
\newcommand*{\xout}{\xoutsym+\arrowLength+2}%

\usetikzlibrary{positioning}
\usetikzlibrary{calc}
\usetikzlibrary{fit}

\tikzset{set/.style={draw,circle,inner sep=0pt,align=center}}

\begin{document}

\title{Symplectic Autoencoders for Model Reduction of Hamiltonian Systems}

\author{
\large{Benedikt Brantner, Michael Kraus}\\
\small{(benedikt.brantner@ipp.mpg.de, michael.kraus@ipp.mpg.de)}
\vspace{.5em}\\
\normalsize{Max-Planck-Institut f\"ur Plasmaphysik}\\
\normalsize{Boltzmannstra\ss{}e 2, 85748 Garching, Deutschland}%
\vspace{.1em}\\
\small{and}
\vspace{.1em}\\
\normalsize{Technische Universit\"at M\"unchen, Zentrum Mathematik}\\
\normalsize{Boltzmannstra\ss{}e 3, 85748 Garching, Deutschland}%
\vspace{1em}\\
}

\date{\today}

\maketitle

\begin{abstract}
	Many applications, such as optimization, uncertainty quantification and inverse problems, require repeatedly performing simulations of large-dimensional physical systems for different choices of parameters. This can be prohibitively expensive. 
	
	In order to save computational cost, one can construct surrogate models by expressing the system in a low-dimensional basis, obtained from training data. This is referred to as model reduction.
	Past investigations have shown that, when performing model reduction of Hamiltonian systems, it is crucial to preserve the symplectic structure associated with the system in order to ensure long-term numerical stability.
	
	Up to this point structure-preserving reductions have largely been limited to linear transformations. We propose a new neural network architecture in the spirit of autoencoders, which are established tools for dimension reduction and feature extraction in data science, to obtain more general mappings.	
	
	In order to train the network, a non-standard gradient descent approach is applied that leverages the differential-geometric structure emerging from the network design.	
	The new architecture is shown to significantly outperform existing designs in accuracy.

\end{abstract}

\newpage

\tableofcontents

\newpage

\section{Introduction}
Real-world physical systems are often described by parametrized partial differential equations (PDEs). In order to solve these, they are ususally approximated by very high-dimensional ordinary differential equations (ODEs). These discretized systems are referred to as \textit{full order model}s (FOM). 
Applications such as optimization, uncertainty quantification and inverse problems require performing repeated simulations of the FOM which can quickly become prohibitively expensive. It has become common practice to construct \textit{reduced order model}s (ROM) in a data-driven way to alleviate this computational bottleneck. Ideally, the ROM is designed to capture the FOM dynamics as accurately as possible (see \cite{antoulas2000survey}). 

ROMs always consist of two mappings: a \textit{reduction} $\mathcal{P}$ from the high- to the low-dimensional space, and a \textit{reconstruction} $\mathcal{R}$ from the low- to the high-dimensional space. Once we have $\mathcal{P}$ and $\mathcal{R}$, we can naturally obtain an ODE on the low-dimensional space by means of \textit{Galerkin projection}. This is ideally much cheaper to solve than the ODE describing the FOM. 

The most-common of these data-driven model reduction techniques is ``proper orthogonal decomposition'' (POD\footnote{This is also known as \textit{principal component analysis} (PCA).}, see \cite{lassila2014model}), in which both $\mathcal{P}$ and $\mathcal{R}$ are orthonormal matrices. POD uses singular value decomposition (SVD) to identify  dominant global modes in the data matrix on which it is performed. 

Despite its ubiquity POD has two limitations: First, it is strictly linear. This means that problems with slowly-decaying singular values (i.e. advection-dominated problems) require a large amount of modes (see \cite{lee2020model, blickhan2023registration}). 

For this reason it is becoming increasingly more common to construct $\mathcal{P}$ and $\mathcal{R}$ using non-linear mappings. If neural networks are used, this amounts to using \textit{autoencoders} (see \cite{lee2020model,fresca2021comprehensive}). Autoencoders are neural networks that represent high-fidelity data using a low-dimensional \textit{feature space}. For reduced order modelling this feature space is the low-dimensional space on which the dynamics evolve.

The second limitation of POD (this is also true for practically all applications of autoencoders) is that it is oblivious to the physical structure of the problem it is applied to. 

This neglect can have grave consequences for the accuracy of the dynamics of the reduced system. An extreme example of this is shown in section \ref{POD_fail}.
Of particular importance among these physical properties is \textit{symplecticity} (see e.g. \cite{arnold1978mathematical}), which is in most cases equivalent to the system being \textit{Hamiltonian}. If the high-fidelity dynamics is Hamiltonian, an alternative to POD exists that takes the Hamiltonian structure of the system into account: Proper Symplectic Decomposition (PSD, \cite{peng2016symplectic,tyranowski2019symplectic}). PSD, however, is affected by the other problem that POD suffers from: It is strictly linear.

In this work we attempt to combine the advantages of autoencoders and PSD by constructing non-linear mappings that also take symplecticity into account (i.e. the Hamiltonian structure of the system).
This is achieved by composing an existing neural network architecture, called \textit{symplectic neural networks} (SympNets, \cite{jin2020sympnets}), with PSD-like matrices. These PSD-like matrices form a manifold. \textit{Training} this new neural network will then require techniques from differential geometry and manifold optimization. 

The remainder of this paper is structured as follows: sections \ref{hamiltonian_systems}, \ref{RBM} and  \ref{sympnets} discuss basic notions of Hamiltonian dynamics, reduced basis methods and SympNets respectively. Section \ref{proposed_architecture} gives a description of the proposed architecture. Section \ref{implementation} discusses details of the implementation and section \ref{experiments} shows results of experiments performed with the new neural networks.  

\section{Hamiltonian Systems}
\label{hamiltonian_systems}

The following offers a brief review of Hamiltonian dynamics. For more details consult e.g. \cite{bishop1980tensor,arnold1978mathematical}. 

A \textit{canonical Hamiltonian system} on $\mathbb{R}^{2N}$ is defined through a Hamiltonian function $H:\mathbb{R}^{2N} \to \mathbb{R}$. The coordinates on $\mathbb{R}^{2N}$ are denoted by $(q^1, \ldots, q^N, p_1, \ldots, p_N)$.

From $H$ we obtain a \textit{Hamiltonian vector field} by first taking the gradient of the function and then applying the matrix\footnote{In theory this can be an arbitrary skew-symmetric non-degenerate matrix. But since one can always find a linear transformation from such a matrix onto $\mathbb{J}_{2N}$ one typically just chooses the latter. Further we will refer to $\mathbb{J}$ as the \textit{symplectic potential}. It is sometimes also called \textit{Poisson tensor}.} $\mathbb{J}_{2N}$:
\begin{equation}
	\mathbb{J}_{2N} = \begin{pmatrix} \mathbb{O}_N & \mathbb{I}_N \\ -\mathbb{I}_N & \mathbb{O}_N \end{pmatrix}.	
\end{equation}

The Hamiltonian vector field hence takes the following form: 
\begin{equation}
	V_H(q, p) = \mathbb{J}_{2N}\nabla{}H(q,p),
	\label{eq:ham_vf}
\end{equation}

or equivalently: 
\begin{equation}
	\dot{q} = \frac{\partial{}H}{\partial{}p}, \quad \dot{p} = -\frac{\partial{}H}{\partial{}q}.
\end{equation}

The flow $\varphi^H_t(\cdot)$ of this differential equation has two very important properties: it preserves the Hamiltonian $H$ and the symplectic potential $\mathbb{J}_{2N}$, i.e. 
\begin{enumerate}
	\item $H\circ\varphi^H_t(z) = H(z)$ and 
	\item $(\nabla\varphi^H_t(z))^T\mathbb{J}_{2N}\nabla\varphi^H_t(z)=\mathbb{J}_{2N}$.
\end{enumerate}
There are both theoretical and empirical arguments showing that considering this structure of the differential equations is extremely important when treating Hamiltonian systems, because it greatly restricts the space of possible solutions (see \cite{gromov1985pseudo, mcduff2017introduction} for details on how symplecticity restricts the space of possible mappings). It has proven very important to consider this structure when designing numerical schemes (see e.g. \cite{hairer2006geometric, sanz2018numerical, leimkuhler2004simulating}).

\subsection{Infinite-dimensional Hamiltonian systems}

Analogous to Hamiltonian ODEs, we can also formulate infinite-dimensional canonical Hamiltonian systems that are PDEs. Given a domain $\Omega$ we define two (time-dependent) functions on it: $q:\mathcal{I}\times\Omega\to\mathbb{R}$ and $p:\mathcal{I}\times\Omega\to\mathbb{R}$. Here a Hamiltonian is a functional that takes $q$ and $p$ as input: 
\begin{equation}
	\mathcal{H}:C^\infty(\mathcal{I}\times\Omega)\times{}C^\infty(\mathcal{I}\times\Omega)\to\mathbb{R}.
\end{equation}

The resulting PDE takes a similar form to the case when we deal with Hamiltonian ODEs:
\begin{equation}
	\partial_t{}q(t,\xi;\mu) = \frac{\delta{}\mathcal{H}}{\delta{}p}, \quad \partial_t{}p(t,\xi;\mu) = -\frac{\delta{}\mathcal{H}}{\delta{}q}.
\end{equation}

An example of this is the ``linear wave equation'' (see \cite{peng2016symplectic, doi:10.1137/21M1466657}) with Hamiltonian 
\begin{equation} 
	\mathcal{H}(q, p; \mu) := \frac{1}{2}\int_\Omega\mu^2(\partial_\xi{}q(t,\xi;\mu))^2 + p(t,\xi;\mu)^2d\xi.
	\label{eq:linear_wave}
\end{equation}

If we discretize the Hamiltonian directly with e.g. a structure-preserving finite difference discretization we again end up with a finite-dimensional Hamiltonian system (but typically in very high dimensions since it is the discretization of a PDE). See section \ref{experiments} for an example of this.

\section{Reduced Basis Methods}
\label{RBM}

The following gives a quick review of data-driven reduced basis methods (RBMs). RBMs can alleviate the cost of solving parametric PDEs (PPDEs) by leveraging machine learning techniques. The task here is the following: 

Assume we are given a PPDE 
\begin{equation}
	F(u(\mu);\mu) = 0, \quad u(\mu)\in{}V,
\end{equation}

where $V$ is some infinite-dimensional Banach space. For solving this PDE we have to first discretize $V$ in some form (finite element, finite volume, finite difference, particle method, \ldots); we denote this discretized space by $V_h$ and refer to the induced differential equation on it as the \textit{full order model} (FOM). In order to achieve sensible simulation results we usually have to choose the dimension of $V_h$ very large, which makes solving the discretized PPDE very expensive. In practice, we typically also have to solve the PPDE for many different parameter instances $\mu$, which can lead to prohibitively high computational costs. Even though the solutions for different parameter instances $\mu\in\mathbb{P}$ will be qualitatively similar, we usually have to run the simulation from scratch every time we change the parameters. Dealing with data and qualitatively similar solutions (also backed up by the concept of the solution manifold\footnote{The solution manifold is the space on which the solutions of the PDE evolve: $\mathcal{M}:=\{u(\mu):\mu\in\mathbb{P}\}$.}) this is an ideal setting to apply machine learning techniques.  

RBMs leverage the parametric structure of the system by constructing a low-dimensional manifold and a corresponding embedding of the manifold in $V_h$ that approximate the dynamics in the high-dimensional space as accurately as possible \cite{fresca2021comprehensive,lassila2014model}. This manifold and the embedding are determined in a data-driven way based on a set of FOM solutions (this process is explained below). In the following, the dimension of the big system is denoted by $2N$ and that of the small system by $2n$. The small system is referred to as the \textit{reduced order model} (ROM).

The ODE in the high-dimensional space $V_h=\mathbb{R}^{2N}$, i.e. the FOM, is denoted by:
\begin{equation}
	\begin{cases}
		\dot{\mathbf{u}}(t;\mu) = \mathbf{f}(t,\mathbf{u}(t;\mu);\mu),\qquad t\in(0,T), \\
		\mathbf{u}(0;\mu) = \mathbf{u}_0(\mu),
	\end{cases}
	\label{eq:FOM}
\end{equation}
where $\mu$ is a set of parameters (potentially also including initial conditions) and $\mathbf{u}(\cdot;\mu)$ is the solution of the ODE for parameters $\mu$ that evolves on $V_h = \mathbb{R}^{2N}$. 

The goal of reduced basis methods is the construction of a reduction $\red:\mathbb{R}^{2N}\to\mathbb{R}^{2n}$ and a reconstruction (or lift) $\rec:\mathbb{R}^{2n}\to\mathbb{R}^{2N}$. The low-dimensional manifold embedded in $V_h$, which was mentioned before, is $\rec(\mathbb{R}^{2n})$ in this case.

In practice, we want to construct $\red$ and $\rec$ based on a set of previous FOM simulations to make successive simulations cheaper.  The data coming from these previous simulations is represented in form of a snapshot matrix 
\begin{equation}
M = [\mathbf{u}(t_0;\mu_i),\ldots,\mathbf{u}(t_T;\mu_i)]_{i=1,\ldots,n_\mathbb{P}}, 
\end{equation} 

where $t_0$ and $t_f$ are the initial and final time of the simulation. With this the problem can be cast into the following form: 
\begin{equation}
	\min_{\red,\rec}d(M,\rec\circ\red(M)),
	\label{eq:reconstruction_error}
\end{equation}
where $d(\cdot,\cdot)$ is an appropriate metric (this can in principle be any metric, but will be $L_2$ from now on).  We refer to equation \eqref{eq:reconstruction_error} as the \textit{reconstruction error}.

Having found the mappings \red{} and \rec{} one has two choices for a vector field on the reduced space:
\begin{align}
	\mathbf{f}^\mathrm{red}(t, z; \mu) & = (\nabla_{\rec(z)}\red)(\mathbf{f}(t, \rec(z); \mu)) \text{ and } \\
	\mathbf{f}^\mathrm{red}(t, z; \mu) & = (\nabla_{z}\rec)^+(\mathbf{f}(t, \rec(z); \mu)), 
	\label{eq:reduced_vector_field}
\end{align}
where $(\cdot)^+$ indicates the pseudo-inverse. The second one of these equations is a ``manifold Galerkin projection'' (see \cite{lee2020model}) and gives the \textit{most consistent} vector field. We will therefore use the second equation to get reduced dynamics from now on. 

For \textit{Hamiltonian model reduction}, both of these mappings have to be symplectic in the sense of Section~\ref{hamiltonian_systems}, i.e., $(\nabla_{(q,p)}\red)\mathbb{J}_{2N}(\nabla_{(q,p)}\red)^T = \mathbb{J}_{2n}$ and $(\nabla_{(\tilde{q},\tilde{p})}\rec)^T\mathbb{J}_{2N}(\nabla_{(\tilde{q},\tilde{p})}\rec) = \mathbb{J}_{2n}$ where $(q, p)$ are the coordinates in the \textit{full space} $\mathbb{R}^{2N}$ and $(\tilde{q}, \tilde{p})$ are the coordinates in the \textit{reduced space} $\mathbb{R}^{2n}$.

\subsection{POD/PSD}

The most widely-used model order reduction technique is Proper Orthogonal Decomposition (POD, see \cite{lassila2014model}). This provides the ideal solution for the case when \red{} and \rec{} are assumed to be \textit{orthonormal} matrices\footnote{More precisely: \rec{} is assumed orthonormal and \textit{the transpose of} \red{} is assumed to be orthonormal, i.e. $\rec=A$ and $\red=B^T$ with $A^TA=B^TB = \mathbb{I}_{2n} \Longleftrightarrow A, B\in{}St(2n, 2N)$.} and an $L_2$ norm is used. That a matrix is orthonormal here means that $A^TA=\mathbb{I}_{2n}$. These orthonormal matrices form a manifold, called the ``Stiefel manifold'' (see \cite{hairer2006geometric, brantner2023generalizing}). The solution then just requires the computation of an SVD for $M$: $M = U\Sigma{}V$. For the POD we set $\mathcal{P}:\mathbb{R}^N\to\mathbb{R}^n$ and $\rec:\mathbb{R}^n\to\mathbb{R}^N$.

\begin{proposition}
	If $\rec \equiv A$ is assumed to be a matrix and $\red = \rec^T$, then $A$ is necessarily orthonormal. It is given by the first $n$ columns of the matrix $U$ in the SVD of $M$.
	\label{th:pod}
\end{proposition}
\begin{proof}
	Consider the SVD $M = U\Sigma{}V^T$. Then the reconstruction error reads $|| M - \rec\circ\red(M) ||_2 = ||U\Sigma{}V^T - AA^TU\Sigma{}V^T||_2 = || (I - AA^T)U\Sigma{}V^T ||_2$. Here $(I - AA^T)$ is a matrix with rank $N-n$ (assuming that $A$ has full rank). The minimum is achieved when $(I - AA^T)$ removes the part of $U$ that contributes the most to the error, i.e. the one for which $\Sigma$ has the highest eigenvalues; this implies $A = [u_1, \ldots, u_n]$, i.e. the first $n$ columns of $U$.
\end{proof}

POD is hence just SVD applied to a snapshot matrix. If the system, represented by $M$, has additional structure (e.g. symplecticity), POD will be entirely oblivious to this. Proper Symplectic Decomposition (PSD, \cite{peng2016symplectic}), on the other hand, encodes the symplectic structure of the system into the reduced basis. In this case the pseudo inverse of $A=\rec$ is $A^+$, the symplectic inverse: $A^+:=\mathbb{J}_{2n}A^T\mathbb{J}_{2N}^T$. This yields the optimization problem:
\begin{equation}
	\min_{A\in{}Sp(2n,2N)}|| M - AA^+M ||_2
\end{equation}
where we look for the minimum on the symplectic Stiefel manifold $Sp(2n,2N) := \{A\in\mathbb{R}^{2N\times2n}:A^T\mathbb{J}_{2N}A = \mathbb{J}_{2n}\}$.

Here we cannot perform an SVD (or a symplectic analog) in the way it was used in the proof of Theorem~\ref{th:pod}, which means we have to rely on techniques like (Riemannian) optimization to find a global minimum on $Sp(2n, 2N)$\footnote{Technically speaking there is an equivalent of SVD where an arbitrary matrix admits a decomposition $M = QDS^{-1}$ where $Q$ is orthogonal and $S$ is symplectic (see \cite{XU20031}). This decomposition can, however, not be used in the way it was done in the proof of Theorem~\ref{th:pod} for SVD.}. Some details of optimizing on manifolds will be discussed in Section~\ref{implementation}. Associated with optimization on $Sp(2n, 2N)$ is another difficulty: even though in principle one could find a minimum of this optimization problem, this is numerically highly unstable as the symplectic group and the symplectic Stiefel manifold are, other than the orthonormal group and the Stiefel manifold, non-compact.

A potential workaround is enforcing an additional constraint such that $A$ is a \textit{cotangent lift}. The matrices we optimize over then take the following form: 
\begin{equation}
	A = \begin{pmatrix} \Phi & \mathbb{O}_{N\times{}n} \\ \mathbb{O}_{N\times{}n} & \Phi  \end{pmatrix}, \quad \Phi\in{St}(n,N).
\label{eq:PSD}
\end{equation}

These matrices are easily verified to lie in $Sp(2n,2N)\cap{}St(2n,2N)$. Computing matrices of the form as in Equation~\eqref{eq:PSD} again just requires an application of SVD and this is referred to as \textit{proper symplectic decomposition} (PSD\footnote{Also not that \textit{PSD-like matrices} shown in equation \eqref{eq:PSD} are naturally isomorphic to the Stiefel manifold $St(n,N)$.}, see \cite{peng2016symplectic}). So the difference between POD and PSD is that for POD we find the minimum of the reconstruction error on all orthonormal matrices, whereas for PSD we find it on matrices as in Equation~\eqref{eq:PSD}; i.e. symplecticity is encoded into the structure. Different approaches to construct elements from $Sp(2n,2N)$, not just the cotangent lift, to minimize the reconstruction error (see Equation~\eqref{eq:reconstruction_error}) are discussed in \cite{gao2022optimization}. In this reference, it is also observed that the cotangent lift works remarkably well, also when compared to different, more involved techniques. Henceforth we will use the terms \textit{cotangent lift} and \textit{PSD} interchangeably.

\subsubsection{Induced differential equations}
\label{posd_eq}

Both POD and PSD induce differential equations on the reduced system through equation \eqref{eq:reduced_vector_field}.
In the case of POD, this is $\dot{z} = A^TV(Az)$ and for PSD it is $\dot{z} = A^+V(Az)$, where $z$ are the reduced coordinates and $V$ is the vector field. Here, $A^+ := \mathbb{J}_{2n}A^T\mathbb{J}_{2N}^T$ is the symplectic inverse that was mentioned above.

If $V\equiv{}V_H$ is a Hamiltonian vector field, then the reduced system in the PSD case will also be Hamiltonian: 
\begin{equation}
	\dot{z} = A^+V_H(Az) = \mathbb{J}_{2n}A^T\mathbb{J}_{2N}^T\mathbb{J}_{2N}\nabla{}H(Az) = \mathbb{J}_{2n}A^T\nabla{}H(Az),
\end{equation}
where the Hamiltonian in the reduced system is now simply $z\mapsto{}H(Az)$.

\subsubsection{Example of failing POD}
\label{POD_fail}

An example where POD can go horribly wrong is the simple Hamiltonian system:
\begin{equation}
	H(\mathbf{q},\mathbf{p}) = \sum_{i=1}^2\frac{p_i^2}{2} + \phi_i\frac{q_i^2}{2},
\end{equation}
with $\phi_1 = 0.05$, $\phi_2 = \pi$ and $((\mathbf{q}^{(0)})^T,(\mathbf{p}^{(0)})^T) = (q^{(0)}_1,q^{(0)}_2,p^{(0)}_1,p_2^{(0)}) = (0,0,1,3)$. This is a system of two independent harmonic oscillators whose frequencies are in an irrational relation to each other; the resulting orbit will thus wind densely around an invariant 2-torus that is embedded in $\mathbb{R}^4$.

If one now takes the snapshot matrix 
\begin{equation}
M := \begin{bmatrix} \mathbf{q}^{(0)} & \mathbf{q}^{(1)} & \cdots & \mathbf{q}^{(T)} \\ 
					 \mathbf{p}^{(0)} & \mathbf{p}^{(1)} & \cdots & \mathbf{p}^{(T)} \end{bmatrix},
\end{equation}
and reduces the system to $\mathbb{R}^2$ (i.e. $N=2$ and $n=1$), then the lifts/reductions have the following form\footnote{These are the matrices obtained in the limit for $hT\to\infty$ and $h\to0$, where $h$ is the stepsize.}: 
\begin{equation}
A_\mathrm{POD} = \begin{pmatrix} 1 & 0 \\ 0 & 0 \\ 0 & 0 \\ 0 & 1 \end{pmatrix}
\quad \text{ and } \quad
A_\mathrm{PSD} = \begin{pmatrix} 1 & 0 \\ 0 & 0 \\ 0 & 1 \\ 0 & 0 \end{pmatrix}. 
\end{equation}
This is because the $L_2$ distance between $(q_1,0,0,p_2)$ and $(q_1,q_2,p_1,p_2)$ is the smallest possible one if only two modes are considered.
For the PSD, the ideal ``reduced system'' is $(q_1,0,p_1,0)$, i.e. the first one of the two circles on the torus $\mathbb{T}^2 \simeq \mathbb{S}^1\times\mathbb{S}^1$.

To make this more clear, consider the full-order dynamics: 
\begin{equation}
	\begin{pmatrix} \mathbf{q}^{(t)} \\ \mathbf{p}^{(t)} \end{pmatrix} = \begin{pmatrix} q_1^{(t)} \\ q_2^{(t)} \\ p_1^{(t)} \\ p_2^{(t)}  \end{pmatrix} = \begin{pmatrix} p_1^{(0)}\sin(\sqrt{\phi_1}t)/\sqrt{\phi_1} \\p_2^{(0)}\sin(\sqrt{\phi_2}t)/\sqrt{\phi_2} \\ p_1^{(0)}\cos(\sqrt{\phi_1}t) \\p_2^{(0)}\cos(\sqrt{\phi_2}t)\end{pmatrix}.
\end{equation}
The ``dominant'' of these dimensions is the first one, since its prefactor is $p_1^{(0)}/\sqrt{\phi_1} = 1/\sqrt{0.05} \approx 4.47$; the second biggest is the forth dimension with prefactor $p_2^{(0)} = 3$.

\begin{figure}
	\begin{subfigure}[b]{.45\textwidth}
	\includegraphics[width=1\textwidth]{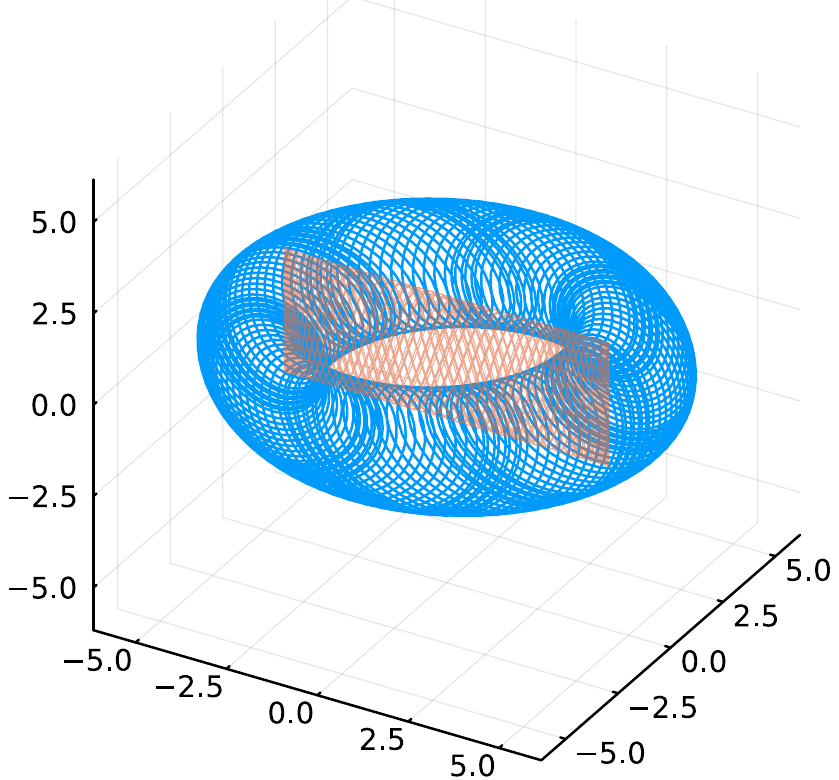}
	\caption{POD: full order model (blue); reduced dynamics (orange).}
	\label{fig:torus_pod}
	\end{subfigure}\hspace{.7cm}\begin{subfigure}[b]{.45\textwidth}
	\includegraphics[width=1\textwidth]{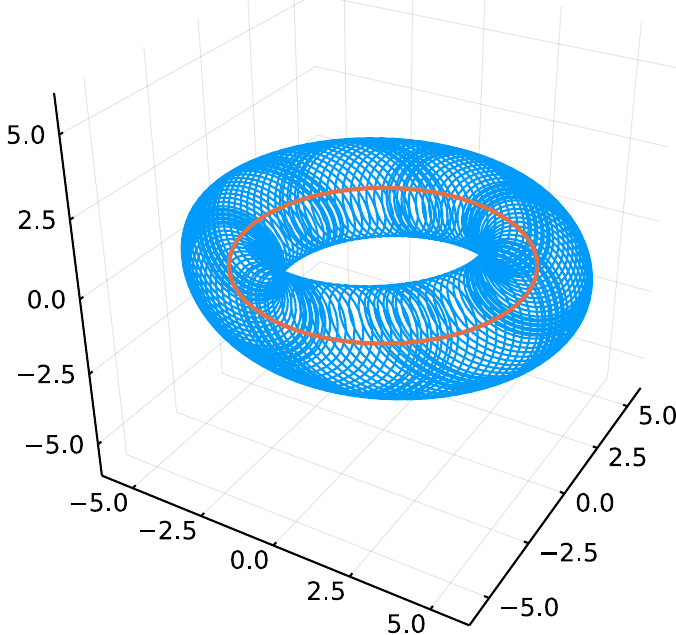}
	\caption{PSD: full order model (blue); reduced dynamics (orange).}
	\label{fig:torus_psd}
	\end{subfigure}
\end{figure}

Since a 2-torus can be embedded in $\mathbb{R}^3$, a visualization of the two reductions is possible. Figures \ref{fig:torus_pod} and \ref{fig:torus_psd} show $A_\mathrm{POD}A_\mathrm{POD}^TM$ and $A_\mathrm{PSD}A_\mathrm{PSD}^+M$ respectively (where $M$ is the data matrix). The embedding for the two plots $\mathbb{T}^2 \hookrightarrow \mathbb{R}^3$ is $(q_1,q_2,p_1,p_2) \mapsto (q_1 + q_2,p_1,p_2)$.

In Figure~\ref{fig:torus_pod}, the POD picks the ``wrong'' two modes, i.e., they are not corresponding $q$ and $p$ coordinates but form a Lissajous curve (orange) in the $xz$-plane instead of a circle. 

The difference between POD and PSD becomes even more apparent when looking at the induced dynamics. The differential equation for the POD is:
\begin{equation} 
	A_\mathrm{POD}^TV_H(A_\mathrm{POD}z) = \begin{pmatrix} 1 & 0 & 0 & 0 \\ 0 & 0 & 0 & 1  \end{pmatrix} \begin{pmatrix} 0 \\ p_2 \\ -\phi_1q_1 \\ 0  \end{pmatrix} = \begin{pmatrix}0 \\ 0 \end{pmatrix}, 
\end{equation}
i.e. the induced trajectory is the constant one. On the other hand, for the PSD, this is: 
\begin{equation}
	A_\mathrm{PSD}^+V_H(A_\mathrm{PSD}z) = \begin{pmatrix} 1 & 0 & 0 & 0 \\ 0 & 0 & 1 & 0 \end{pmatrix} \begin{pmatrix}  p_1 \\ 0 \\ -\phi_1q_1 \\ 0  \end{pmatrix} = \begin{pmatrix} p_1 \\ -\phi_1q_1 \end{pmatrix},
\end{equation}
i.e. the dynamics are correctly reduced to the first one of the two harmonic oscillators. Another example of a failing POD for the Vlasov equation is shown in \cite{tyranowski2019symplectic}. 

\subsection{Autoencoders}

Whereas the previous section addressed that POD does not encode the Hamiltonian dynamics in its reduced model, this section will deal with how to remove the constraint that it is strictly linear.

The term ``autoencoder'' refers to a pair of neural networks, the encoder $\red = \Psi^\mathrm{enc}$ and the decoder $\rec = \Psi^\mathrm{dec}$, that are ``trained'' to minimize a reconstruction error $d(M,\Psi^\mathrm{dec}\circ\Psi^\mathrm{enc}(M))$ for a given dataset $M$ (see \cite[Chapter 14]{goodfellow2016deep}). If $\Psi^\mathrm{enc}$ and $\Psi^\mathrm{dec}$ are assumed to be linear (i.e. matrices), one again arrives at POD, which may therefore be viewed as a simple, linear form of an autoencoder.
Originally, autoencoders were devised to perform feature extraction, but have also successfully been applied to image generation (see \cite{bank2020autoencoders}). The following gives a short review of feedforward neural networks, the simplest instance of a general neural network that can be used to construct autoencoders. This will be used to introduce our neural network architecture in Section~\ref{proposed_architecture}.

\subsubsection{Feedforward Neural Networks}

Feedforward neural networks offer a method to build a basis for general functions. In e.g. \cite{hornik1989multilayer} it is shown that feedforward neural networks can approximate any continuous function in a certain topology: 

\begin{theorem}[Universal approximation theorem for feedforward neural networks \cite{hornik1989multilayer}]
	The set of two-layer neural networks is uniformly dense on compacta in $C^0(\mathbb{R}^N)$, the set of continuous functions $\mathbb{R}^N\to\mathbb{R}$ if the activation function is non-polynomial.
	\label{th:univ_approx}
\end{theorem}

The theorem, of course, also holds for neural networks that have more than two layers. A layer of a feedforward neural network consists of an application of an affine transformation, i.e., $x\mapsto Ax + b$ (where $A\in\mathbb{R}^{m\times{}n}$ and $b\in\mathbb{R}^m$) and a \textit{nonlinearity}\footnote{Popular choices for nonlinearities include the \textit{nonlinear rectified unit} (RELU) $x\mapsto\mathrm{max}(0,x)$ and $\mathrm{tanh}(x) = (1 - e^{-x})/(1 + e^{-x})$.} $\sigma$.
The last layer of a feedforward neural network is typically just a linear transformation without the nonlinearity. Theorem~\ref{th:univ_approx} states that the mapping $x \mapsto v^T\sigma(Ax +b)$ (with $v\in\mathbb{R}^m$) can approximate any function (on a compact set) if $m$ is chosen big enough. 

The problem of finding the parameters $A$, $b$ and $v$ that parametrize the function is usually solved in an iterative manner that is referred to as the \textit{training} of the neural network. 
The training aims to minimize a \textit{loss function} that maps the weights $\theta:=(A,b,v)$ to the non-negative real numbers $\mathbb{R}^+ := [0,\infty)$. For the autoencoder example with feedforward neural networks as encoder and decoder, the loss function is the reconstruction error: 
\begin{equation}
	L: \theta \equiv (\theta_1,\theta_2) \mapsto d(M,\Psi^\mathrm{dec}_{\theta_1}\circ\Psi^\mathrm{enc}_{\theta_2}(M)). 
\end{equation}
The simplest form of training is gradient descent with a constant learning rate $\eta$, where the optimization is performed iteratively: $\theta^{i+1}\leftarrow \theta^i - \eta\nabla_{\theta^i}L$. More involved optimization techniques are discussed in Section~\ref{implementation}. 

Section~\ref{sympnets} gives a brief discussion of \textit{SympNets}. These are a special type of feedforward neural network that can approximate any symplectic map. To this end the space of maps that a neural network can approximate is restricted to symplectic maps.

\section{SympNets}
\label{sympnets}

SympNets (see \cite{jin2020sympnets}) offer a way of constructing general symplectic maps with simple building blocks. The collections of networks which can be constructed with these building blocks are called  $\Psi_{LA}$ (sympnets made up of \textit{linear} and \textit{activation} modules) and $\Psi_G$ (sympnets made up of \textit{gradient} modules). $\Psi_{LA}$ and $\Psi_G$ are dense in $\mathcal{SP}^r(\mathbb{R}^{2N})$, the set of $r$-differentiable symplectic maps (for arbitrary $r$), in some topology (similarly to theorem \ref{th:univ_approx}). More concretely: 

\begin{theorem}[Universal approximation theorem for SympNets \cite{jin2020sympnets}]
	For any $r\in\mathbb{N^*}$, $\Psi_{LA}$ and $\Psi_G$ are both $r$-uniformly dense on compacta in $\mathcal{SP}^r(\mathbb{R}^{2N})$ if the activation function $\sigma$ is $r$-finite. 
\end{theorem}

$\sigma$ being $r$-finite means that $0<\int|D^r(\sigma(x))|dx<\infty$. $r$-uniformly dense on compacta means that for each $g\in\mathcal{SP}^r(\mathbb{R}^{2N})$, $K$ compact, $\epsilon>0$ there exists $\psi_{g,K,\epsilon}\in\Psi_{LA}/\Psi_G$ such that $||g -\psi_{g,K,\epsilon}||_r < \epsilon$, where $||\cdot||_r$ is the Sobolev norm.  

The components making up $\Psi_{LA}$ and $\Psi_G$ will now briefly be reviewed. 

\subsection{Linear/Activation-type SympNets $\Psi_{LA}$}
\textit{Linear SympNet layers} consist of upper/lower triangular matrices of the following form: 

\begin{equation}
	\begin{pmatrix}\mathbb{I}_N & S \\
		\mathbb{O}_N & \mathbb{I}_N
	\end{pmatrix}\text{ and }
	\begin{pmatrix}\mathbb{I}_N & \mathbb{O}_N \\ 
		S' & \mathbb{I}_N
	\end{pmatrix},
	\label{eq:linear}
\end{equation}

where $S$ and $S'$ are symmetric matrices. These matrices are by construction symplectic. 

In \cite{jin2022optimal} it was shown that every symplectic matrix (i.e. an element of $Sp(2N,2N)\equiv{}Sp(2N)$) can be represented by 5 of these linear layers. 

\textit{Activation layers} form the remaining component(s) of $\Psi_{LA}$. They take the following form:

\begin{equation}
	\begin{pmatrix} q \\ p \end{pmatrix} \mapsto \begin{pmatrix} q + \mathrm{diag}(a)\sigma(p) \\ p \end{pmatrix} \text{ and } \begin{pmatrix} q \\ p \end{pmatrix} \mapsto \begin{pmatrix} q \\ p + \mathrm{diag}(a')\sigma(q) \end{pmatrix}, 
		\label{eq:activation}
\end{equation}

where $a\in\mathbb{R}^n$ and $[\mathrm{diag}(a)]_{ij} = a_i\delta_{ij}$. $\Psi_{LA}$ is then the set of all possible combinations of the layers presented in equations \eqref{eq:linear} and \eqref{eq:activation}. 

\subsection{Gradient-type SympNets $\Psi_G$}
The third type of SympNet layers are \textit{gradient layers}. They have the following form: 

\begin{equation}
	\begin{pmatrix} q \\ p  \end{pmatrix} \mapsto \begin{pmatrix} q + K^T\mathrm{diag}(a)\sigma(Kp + b) \\ p  \end{pmatrix} \text{ and } 	\begin{pmatrix} q \\ p  \end{pmatrix} \mapsto \begin{pmatrix} q \\ p + K'^T\mathrm{diag}(a')\sigma(K'q + b)  \end{pmatrix},
		\label{eq:gradient}
\end{equation}
where $K\in\mathbb{R}^{L\times{}N}$ and $a,b\in\mathbb{R}^{L}$ ith $L$ arbitrary. $\Psi_G$ is any combination of layers of the type presented in equation \eqref{eq:gradient}.

The \textit{symplectic autoencoders} presented in this work use SympNets and PSD-like matrices to construct mappings that preserve symplecticity and also change dimension. The first mapping in equation \eqref{eq:gradient} will be refered to as \texttt{GradientQ} and the second as \texttt{GradientP} as they change the $q$ and the $p$ component respectively.

\section{Proposed Architecture}
\label{proposed_architecture}

The following illustrates the proposed architecture:

{

\centering

\begin{tikzpicture}[scale=0.5]

    \begin{scope}
        \foreach \i in {\xMin,...,\xMax} {
            \draw [rectangle, draw, fill=\colIn] (\xin,\xMax/2-\i) rectangle (\xin+1,\xMax/2-\i+1) node [pos=.5] {$z_\i$};
        }
    \end{scope}

    \path [draw,->] (\xin+1.5,0) -- (\xinsym-0.5,0) node [pos=0.5,above,align=left] {$\psi_1$};

    \begin{scope}
        \foreach \i in {\xMin,...,\xMax} {
            \draw [rectangle, draw, fill=\colSym] (\xinsym,\xMax/2-\i) rectangle (\xinsym+1,\xMax/2-\i+1) node [pos=.5] {};
        }
    \end{scope}

    \path [draw,->] (\xinsym+1.5,0) -- (\xenc-0.5,0) node [pos=0.5,above,align=left] {$A_1$};

    \begin{scope}
        \foreach \i in {\xMin,...,\xMaxEnc} {
            \draw [rectangle, draw, fill=\colEnc] (\xenc,\xMaxEnc/2-\i) rectangle (\xenc+1,\xMaxEnc/2-\i+1) node {};
        }
    \end{scope}

    \path [draw,->] (\xenc+1.5,0) -- (\xencsym-0.5,0) node [pos=0.5,above,align=left] {$\psi_2$};

    \begin{scope}
        \foreach \i in {\xMin,...,\xMaxEnc} {
            \draw [rectangle, draw, fill=\colSym] (\xencsym,\xMaxEnc/2-\i) rectangle (\xencsym+1,\xMaxEnc/2-\i+1) node {};
        }
    \end{scope}

    \path [draw,->] (\xencsym+1.5,0) -- (\xintr-0.5,0) node [pos=0.5,above,align=left] {$A_2$};

    \begin{scope}
        \foreach \i in {\xMin,...,\xMaxIntr} {
            \draw [rectangle, draw, fill=\colIntr] (\xintr,\xMaxIntr/2-\i) rectangle (\xintr+1,\xMaxIntr/2-\i+1) node {};
        }
    \end{scope}

    \path [draw,->] (\xintr+1.5,0) -- (\xdecsym-0.5,0) node [pos=0.5,above,align=left] {$A_3^+$};

    \begin{scope}
        \foreach \i in {\xMin,...,\xMaxEnc} {
            \draw [rectangle, draw, fill=\colSym] (\xdecsym,\xMaxEnc/2-\i) rectangle (\xdecsym+1,\xMaxEnc/2-\i+1) node {};
        }
    \end{scope}

    \path [draw,->] (\xdecsym+1.5,0) -- (\xdec-0.5,0) node [pos=0.5,above,align=left] {$\psi_3$};

    \begin{scope}
        \foreach \i in {\xMin,...,\xMaxDec} {
            \draw [rectangle, draw, fill=\colDec] (\xdec,\xMaxDec/2-\i) rectangle (\xdec+1,\xMaxDec/2-\i+1) node {};
        }
    \end{scope}

    \path [draw,->] (\xdec+1.5,0) -- (\xoutsym-0.5,0) node [pos=0.5,above,align=left] {$A_4^+$};

    \begin{scope}
        \foreach \i in {\xMin,...,\xMax} {
            \draw [rectangle, draw, fill=\colSym] (\xoutsym,\xMax/2-\i) rectangle (\xoutsym+1,\xMax/2-\i+1) node [pos=.5] {};
        }
    \end{scope}

    \path [draw,->] (\xoutsym+1.5,0) -- (\xout-0.5,0) node [pos=0.5,above,align=left] {$\psi_4$};

    \begin{scope}
        \foreach \i in {\xMin,...,\xMax} {
            \draw [rectangle, draw, fill=\colOut] (\xout,\xMax/2-\i) rectangle (\xout+1,\xMax/2-\i+1) node [pos=.5] {$\tilde{z}_\i$};
        }
    \end{scope}

\end{tikzpicture}

}

It is a composition of SympNets and PSD-like matrices:
\begin{align}
	\Psi^\mathrm{enc} & = \psi^L_\mathrm{symp}\circ(A_{PSD}^L)^+\circ\cdots\circ\psi^2_\mathrm{symp}\circ(A_{PSD}^1)^+\circ\psi^1_\mathrm{symp}\text{ and} \\ \Psi^\mathrm{dec} & = \tilde{\psi}^{\tilde{L}}_\mathrm{symp}\circ\tilde{A}_{PSD}^{\tilde{L}}\circ\cdots\circ\tilde{\psi}^2_\mathrm{symp}\circ\tilde{A}_{PSD}^1\circ\tilde{\psi}^1_\mathrm{symp}.
\end{align}

The SympNet layers here perform a \textit{symplectic preprocessing} step before the linear, symplectic reduction is employed with the PSD-like layers. The PSD-like layers are matrices of the same form as in equation \eqref{eq:PSD}. 

This then yields a neural network that is capable of approximating any canonical symplectic mapping, also mappings that change the dimension of the system - this is not possible with traditional SympNets. 

One has to be careful with training this network since some of the weights have to satisfy the symplecticity constraint. In order to train the \textit{SympNet part} of the network, standard general-purpose algorithms can be employed as was also done in the original SympNet paper \cite{jin2020sympnets}. Optimizing the elements of the Stiefel manifold(s) is more complicated, as the condition $\Phi^T\Phi = \mathbb{I}_{n}$ has to be enforced in each training step. This naturally leads to \textit{manifold optimization} (see \cite{bendokat2021real, gao2021riemannian, absil2008optimization}).

The Stiefel manifold $St(n, N)\simeq\{\text{$PSD$-like matrices}\}\subset\mathbb{R}^{2N\times2n}$  is a homogeneous space for which a generalization of common optimizers exist (see \cite{brantner2023generalizing}). In addition the set $St(n, N)$ is compact (whereas $Sp(2n, 2N)$, for example, is not); this makes optimization on $St(n, N)$ relatively easy. 

Note that according to equation \eqref{eq:reduced_vector_field}, if the jacobian of the reconstruction $\Psi^\mathrm{dec}$ is symplectic, then one obtains a Hamiltonian system with Hamiltonian $H\circ\Psi^\mathrm{dec}$ on the reduced space. This can be achieved with the new architecture. 

The following section summarizes the most important aspects of optimization on manifolds relevant for neural networks.

\section{Manifold Optimization for Neural Networks}
\label{implementation}

This section discusses mathematical and computational aspects relevant for the implementation of the new neural network architecture. This requires generalizing existing optimizers to the manifolds that appear in the network, i.e. $St(n, N)$. This manifold, as well as $Sp(2n,2N)$, are homogeneous spaces, a special subclass of manifolds that make generalizing common optimizers (stochastic gradient descent with momentum, Adam, etc.) possible. Details regarding this are presented in \cite{brantner2023generalizing}. For the purposes of this paper we limit the discussion to standard first-order optimization on general Riemannian manifolds, i.e. generalized gradient descent, and mention Adam only briefly. 

\subsection{Optimization on Riemannian manifolds and retractions}

Here a generalization of the update rule $Y^{t+1}\leftarrow Y^t - \eta\nabla_{Y^t}L$ from Euclidean spaces to manifolds is discussed. In practice this involves two steps: the computation of a \textit{Riemannian gradient} and a \textit{retraction} that (approximately) solves the geodesic equation.

Assume we are given a Riemannian manifold $(\mathcal{M}, g)$ and an objective function $L$ that we want to minimize on $\mathcal{M}$. Standard \textit{gradient descent} does this by repeatedly solving the \textit{geodesic equation}. This is a canonical second-order ODE associated with all Riemannian manifolds. This ODE is obtained by applying the variational principle (see \cite{kraus2013variational}) to the following functional: 
\begin{equation}
	\mathcal{L}(\gamma) = \int_0^1\sqrt{g_{\gamma(t)}(\dot{\gamma}(t), \dot{\gamma}(t))}.
	\label{eq:geodesic_variation}
\end{equation}

The \textit{geodesic equation}, the result of this variational principle, is then usually written in the following form: 
\begin{equation}
	\frac{d^2\gamma^\lambda}{dt^2} + \Gamma^\lambda_{\mu\nu}\frac{d\gamma^\mu}{dt}\frac{d\gamma^\nu}{dt} = 0,
	\label{eq:geodesic_equation}
\end{equation}

with $\Gamma^\lambda_{\mu\nu}$ begin the \textit{Christoffel symbols of the second kind} (see \cite[Chapter 5.12]{bishop1980tensor}); the initial conditions are $\gamma(0) = Y$ and $\dot{\gamma}(0) = W$. It is worth mentioning that we do not directly work with equation \eqref{eq:geodesic_equation} and Christoffel symbols, but exploit the special structure of the manifolds we deal with (as homogeneous spaces) to make solving the geodesic equation computationally more tractable. For optimizing neural networks with weights on manifolds we now need a value for $W$ (i.e. an initial condition) and a method to solve or approximate equation \eqref{eq:geodesic_equation}.

First we address how to obtain the value for $W$, i.e. the initial condition for the second order ODE. If we are optimizing neural networks with weights on vector spaces, $W$ is the result of automatic differentiation (AD, \cite{innes2019differentiable, moses2021reverse}) multiplied with the negative of a learning rate, i.e. $W = -\eta\nabla{}L$. If we are instead dealing with more general manifolds, we have to perform another operation to obtain an element of the tangent space $W\in{}T_{Y^t}\mathcal{M}$. This additional operation is called computing the \textit{Riemannian gradient}:
\begin{definition}\label{def:riemannian_gradient}
	The \textbf{Riemannian gradient} of a function $L$ on a Riemannian manifold $(\mathcal{M}, g)$ is an element of the tangent space $\mathrm{grad}_YL\in{}T_Y\mathcal{M}$ such that $\forall{}V\in{}T_Y\mathcal{M}:\langle{}dL, V\rangle = g_Y(\mathrm{grad}_YL, V)$.  
\end{definition}

For $St(n, N)$ and $Sp(2n, 2N)$ the Riemannian gradient can be obtained through a simple operation on the output of an AD routine\footnote{The output of the AD routine is called $\nabla_YL$ here, i.e. the Euclidean gradient of $L$ at $Y$.}:
\begin{align}\label{eq:stsp_gradient}
	\mathrm{grad}_YL = & \nabla_{Y}L - Y(\nabla_{Y}L)^TY & \text{for $Y\in$} & St(n, N), & \text{and} \\
	\mathrm{grad}_YL = & \nabla_{Y}L  (Y^T  Y) + \mathbb{J}_{2N}  Y  ((\nabla_{Y}L)^T  \mathbb{J}_{2N}  Y) & \text{for $Y\in$} & Sp(2n, 2N). &
\end{align}

Also see appendix \ref{riemannian_gradient} for a discussion of these expressions and \cite{absil2008optimization, bendokat2021real, gao2021riemannian, kong2022momentum} for similar computations. After having obtained the Riemannian gradient of $L$ at $Y^t$ one can solve or approximate the geodesic equation. For $St(n, N)$ and $Sp(2n, 2N)$ there exist closed-form solutions for the geodesic equation, see \cite[equation (2.25)]{edelman1998geometry} and \cite[equation (3.18)]{bendokat2021real}. Alternatively one could approximate the solution of the geodesic with so-called \textit{retractions}, one of the most popular of these is the \textit{Cayley transform} (see e.g. \cite[chaber IV]{hairer2006geometric} and \cite[chapter 4.1]{absil2008optimization}). The solution of equation \eqref{eq:geodesic_equation} with $\gamma(0) = Y$ and $\dot{\gamma}(0) = V\in{}T_Y\mathcal{M}$ will be called $\gamma_\mathcal{M}(t; V)$.

To sum up, the update rule $Y^{t+1}\leftarrow{}Y^t - \eta\nabla_{Y^t}L$ on regular vector spaces is replaced with:
\begin{equation}
	Y^{t+1} \leftarrow \gamma_\mathcal{M}(1; -\eta\mathrm{grad}_YL).
\end{equation}

So we first compute the \textit{Euclidean gradient} $\nabla_YL$ with an AD routine, then obtain the \textit{Riemannian gradient} with equation \eqref{eq:stsp_gradient} and then use this as an initial condition to solve (or approximate) the geodesic equation in \eqref{eq:geodesic_equation}.

\subsection{Generalizing Adam}

Modern training of neural networks does not simply perform gradient descent, but uses optimization algorithms that include first and second moments (called $s$ and $r$ in the algorithm below), which we also refer to as the \textit{cache}. One of the most successful optimization algorithms is Adam \cite{kingma2014adam}:

\begin{algorithm}[H]
	\begin{algorithmic}[1]
		\Require{step size $\eta$; exponential decay rates $\beta_1,\beta_2\in[0,1)$; constant for numerical stabilization $\delta$; first and second moments $s^{t-1}$, $r^{t-1}$};
		\State compute $G_\theta \leftarrow \nabla_\theta{}L(\theta^{t-1})$,
		\State update $s^t \leftarrow \frac{\beta_1 - \beta_1^t}{1 - \beta_1^t}s^{t-1} + \frac{1-\beta_1}{1-\beta_1^t}G_\theta$,
		\State update $r^t \leftarrow \frac{\beta_2 - \beta_2^t}{1 - \beta_2^t}r^{t-1} + \frac{1-\beta_2}{1-\beta_2^t}G_\theta\odot{}G_\theta$,
		\State update $\theta^{t} \leftarrow \theta^{t-1} - \eta\frac{s^t}{\sqrt{r^t+\delta}}$ \Comment{operations are applied element-wise}.
	\end{algorithmic}
	\caption{Pseudocode for a single gradient step with Adam.}
	\label{alg:adam}
\end{algorithm}

The Adam algorithm (and other neural network optimizers) take as input the current neural network parameters $\theta^{t-1}$ and a \textit{cache} (in the Adam algorithm above this is $s^{t-1}$ and $r^{t-1}$) and updates both of them. Some aspects of this algorithm are easy to generalize to manifolds, other are not. The following offers a short description of this, breaking the alorithm into three main parts:
\begin{enumerate}
\item \textbf{Computing a first order derivative of the \textit{loss function} with respect to the network parameters.} For usual neural network weights this is the output of an AD routine (the Euclidean gradient). This can be easily generalized to manifolds by computing the Riemannian gradient (see equation \eqref{eq:stsp_gradient}).
\item \textbf{Updating the cache and computing a \textit{final velocity}.} In the vector space case this final velocity is $v^t = -\eta{}s^t/(r^t + \delta)$. This is the step that is very difficult to generalize to manifolds since many operations (such as the Hadamard product) in the Adam algorithm lack a clear physical interpretation. 
\item \textbf{Performing the update.} For the vector space case this is simply adding the velocity onto the neural network weights, i.e. $\theta^t\gets\theta^{t-1} + v^t$. For manifolds this step consists of solving (or approximating) the geodesic equation \eqref{eq:geodesic_equation}.
\end{enumerate}

	In geometric terms, step 2 above (i.e. he Hadamard product and associated operations) depends on the chosen coordinate system and is inherently non-geometric. A change of coordinates produces a different result. A workaround for this exists by leveraging the goemetry of \textit{homogeneous spaces}, a category to which the Stiefel manifold and the symplectic Stiefel manifold belong. Working on these we can identify a \textit{global tangent space representation} $\mathfrak{g}^\mathrm{hor}$. This is the space on which the cache is defined and all the Adam operations that lack a clear geometric interpretation are performed, i.e. the operations $\odot$, $\sqrt{(\cdot)}$, $/$ and $(\cdot)+\delta$ in algorithm \ref{alg:adam}. The full generalization of Adam to homogeneous spaces is discussed in \cite{brantner2023generalizing}. This is also the algorithm that we use for training the networks in the following section.

\section{Experiments}
\label{experiments}

We demonstrate the new method on the example of the \textit{linear wave equation} (see equation \eqref{eq:linear_wave}) with $\xi\in\Omega:=(-1/2,1/2)$ and $\mu\in\mathbb{P}:=[5/12,2/3]$ as a choice for domain and parameters (similarly to \cite{doi:10.1137/21M1466657}). 
We additionaly specify the following initial and boundary conditions: 

\vspace{-.7cm}
\begin{align}
	q_0(\omega;\mu) := & q(0, \omega; \mu) \\ 
	p(0, \omega; \mu) = \partial_tq(0,\xi;\mu) = & -\mu\partial_\omega{}q_0(\xi;\mu) \\
	q(t,\omega;\mu) = & 0, \text{ for } \omega\in\partial\Omega.
\end{align}

The precise shape of $q_0(\cdot;\cdot)$ is described below.

As with any other PDE, the wave equation can also be discretized to obtain a ODE which can be solved numerically.

If we discretize $\mathcal{H}$ directly, to obtain a Hamiltonian on a finite-dimensional vector space, we get a Hamiltonian ODE\footnote{This conserves the Hamiltonian structure of the system.}:

\begin{equation}
\mathcal{H}_h(z) = \sum_{i=1}^{\tilde{N}}\frac{\Delta{}x}{2}\bigg[p_i^2 + \mu^2\frac{(q_i - q_{i-1})^2 + (q_{i+1} - q_i)^2}{2\Delta{}x}\bigg] = \frac{\Delta{}x}{2}p^Tp + z^TKz,
\label{eq:discretized_hamiltonian}
\end{equation}

where the components of the matrix $K$ are of the form: 

\begin{equation}
k_{ij} = \begin{cases}  \frac{\mu^2}{4\Delta{}x} &\text{if $(i,j)\in\{(0,0),(\tilde{N}+1,\tilde{N}+1)\}$ }, \\
    -\frac{\mu^2}{2\Delta{}x} & \text{if $(i,j)=(1,0)$ or $(i,j)=(\tilde{N},\tilde{N}+1)$} \\
    \frac{3\mu^2}{4\Delta{}x} & \text{if $(i,j)\in\{(1,1),(\tilde{N},\tilde{N})\}$} \\
    \frac{\mu^2}{\Delta{}x} & \text{if $i=j$ and $i\in\{2,\ldots,(\tilde{N}-2)\}$} \\ 
    -\frac{\mu^2}{2\Delta{}x} & \text{if $|i-j|=1$ and $i,j\notin\{0,\tilde{N}+1\}$} \\
                        0 & \text{else}.
                        \end{cases}
\end{equation}

and $q = (q_0, q_1, \cdots, q_{\tilde{N}})^T,\, p = (p_0, p_1, \cdots, p_{\tilde{N}})^T$.

$\tilde{N}$ is the number of grid points, which leads to a symplectic vector space of dimension $2N := 2(\tilde{N}+2)$.

The vector field of the FOM is described by (see \cite{peng2016symplectic}):

\begin{equation}
  \frac{dz}{dt} = \mathbb{J}_d\nabla_z\mathcal{H}_h = \mathbb{J}_d\begin{bmatrix}\Delta{}x\mathbb{I}  & \mathbb{O} \\ \mathbb{O} & K + K^T\end{bmatrix}z, \quad \mathbb{J}_d = \frac{\mathbb{J}_{2N}}{\Delta{}x}.
	\label{eq:linear_wave_discretized_vector_field}
\end{equation}

The parameter-dependence of this system is encoder into the matrix $K$. Note that, as opposed to equation \eqref{eq:FOM}, this system is not time-dependent. In the following section we describe how we obtain the initial conditions.

\subsection{Initial conditions}

The initial conditions are based on the following third-degree spline (also used in \cite{doi:10.1137/21M1466657}): 

\begin{equation}
h(s)  = \begin{cases}
        1 - \frac{3}{2}s^2 + \frac{3}{4}s^3 & \text{if } 0 \leq s \leq 1 \\ 
        \frac{1}{4}(2 - s)^3 & \text{if } 1 < s \leq 2 \\ 
        0 & \text{else.} 
\end{cases}
\label{eq:spline}
\end{equation}

Plotted on the relevant domain it takes the following shape: 

\begin{figure}[H]
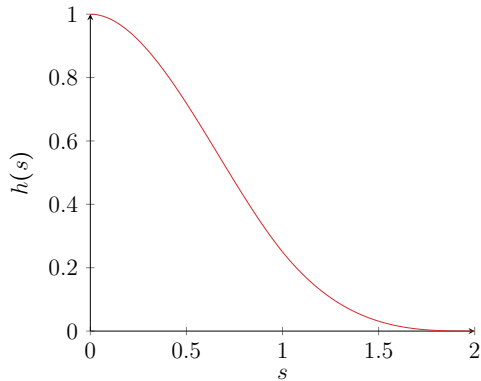

	\centering

	\includestandalone[width=.4\textwidth]{third_degree_spline}
	\caption{The third-degree spline from equation \eqref{eq:spline}. This is used to model all initial conditions.}
	\label{fig:spline}
\end{figure}

Taking the above function $h(s)$ as a starting point, the initial conditions for the linear wave equations are modelled with 

\begin{equation}
	q_0(\omega;\mu) = h(s(\omega, \mu)).
\end{equation}

For the construction of $s(\cdot;\cdot)$ we consider the following factors: 
\begin{enumerate}
\item the analytic solution for equation \eqref{eq:linear_wave} is $q(t, \omega; \mu) = q_0(\omega - \mu{}t; \mu)$; i.e. the solutions of the linear wave equation will travel with speed $\mu$. We should make sure that the wave does not \textit{touch} the right boundary of the domain for any of the values $t\in\mathcal{I}$. So the peak should be sharper for higher values of $\mu$ as the wave will travel faster.
\item the wave should start at the left boundary of the domain in order to cover it as much as possible. 
\end{enumerate}

Based on this we end up with the following choice of parametrized initial conditions: 

\begin{equation}
q_0(\mu)(\omega) = h(s(\omega, \mu)), \quad s(\omega, \mu) =  20 \mu  \left|\omega + \frac{\mu}{2}\right|.
\end{equation}

\begin{figure}

	\centering

	\includegraphics[width=.7\textwidth]{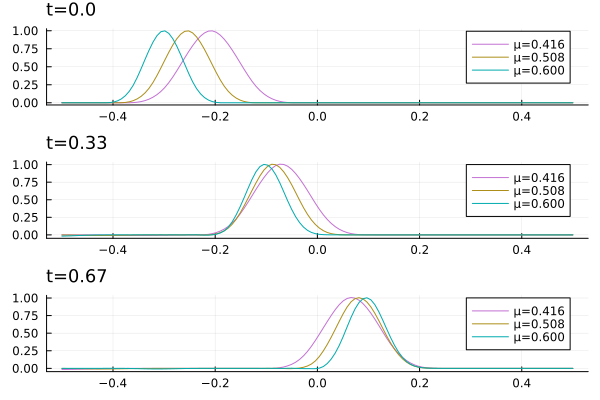}
	\caption{Numerical solutions of equation \eqref{eq:linear_wave} for three different values of $\mu$. Note that the shape of the curves as well as the speed with which they travel changes based on the parameter.}
	\label{fig:waves}
\end{figure}

Three initial conditions and their time evolutions are shown in figure \ref{fig:waves}. As was required, we can see that the peak gets sharper and moves to the left as we increase the parameter $\mu$; the curves also get a good coverage of the domain $\Omega$.

The wave equation has a slowly-decaying Kolmogorov $n$-width (see e.g. \cite{greif2019decay}), which means linear methods like PSD will perform poorly.

For discretizing the system we choose $\tilde{N}$ in equation \eqref{eq:discretized_hamiltonian} to be $\tilde{N}=128$, resulting in a FOM dimension of $2N = 2(\tilde{N}+2) = 260$; here we get two additional degrees of freedom for the $q$ and $p$ coordinates because we also have to consider the end-points (boundary conditions). We integrate the system in equation \eqref{eq:linear_wave_discretized_vector_field} with implicit midpoint, which is a general-purpose symplectic integrator (see \cite{hairer2006geometric}).

\begin{figure}

	\centering

\begin{tikzpicture}[module/.style={draw, very thick, rounded corners, minimum width=15ex},
    gradient/.style={module, fill=mgreen!50},
    psd/.style={module, fill=mblue!50},
    arrow/.style={-stealth, thick, rounded corners},
]
\node (input) {Input};
\node[above of=input, gradient, align=center, yshift=.6cm] (gq1) {\texttt{GradientQ(2N, 10N, tanh)}};
\node[above of=gq1, gradient, align=center] (gq2) {\texttt{GradientP(2N, 10N, tanh)}};
\node[above of=gq2, gradient, align=center] (gq3) {\texttt{GradientQ(2N, 10N, tanh)}};
\node[above of=gq3, gradient, align=center] (gq4) {\texttt{GradientP(2N, 10N, tanh)}};
\node[above of=gq4, psd, align=center] (psd1) {\texttt{PSDLayer(2N, 2n)}};

\node[right of=input, xshift=6cm] (output) {Output}; 
\node[above of=output, gradient, yshift=.6cm] (gq7) {\texttt{GradientQ(2N, 2N, tanh)}};
\node[above of=gq7, psd] (psd2) {\texttt{PSDLayer(2n, 2N)}};
\node[above of=psd2, gradient] (gq6) {\texttt{GradientP(2n, 10n, tanh)}};
\node[above of=gq6, gradient] (gq5) {\texttt{GradientQ(2n, 10n, tanh)}};

\draw[arrow] (gq1) -- (gq2);
\draw[arrow] (gq2) -- (gq3); 
\draw[arrow] (gq3) -- (gq4);
\draw[arrow] (gq4) -- (psd1); 

\coordinate[above of=psd1] (psd_above);
\coordinate[above of=gq5] (gq5_above);

\draw[arrow] (gq5) -- (gq6);
\draw[arrow] (gq6) -- (psd2);
\draw[arrow] (psd2) -- (gq7);

\node[fit=(gq1)(gq2)(gq3)(gq4)(psd1),draw, ultra thick, rounded corners, label=left:$\mathcal{P}$] (encoder) {};
\node[fit=(gq5)(gq5)(gq7)(psd2),draw, ultra thick, rounded corners, label=left:$\mathcal{R}$] (decoder) {};

\draw[arrow] (encoder.north)-|(psd_above)-|(gq5_above)--(decoder.north);
\draw[arrow] (input) -- (encoder);
\draw[arrow] (decoder) -- (output);

\end{tikzpicture}
\caption{The architecture we use for all experiments. \texttt{GradientQ} and \texttt{GradientP} are the gradient layers (see section \ref{sympnets}) that change the $q$ and $p$ component respectively. The syntax is the same used as in \texttt{GeometricMachineLearning.jl}.}
\label{fig:encoder_decoder}
\end{figure}
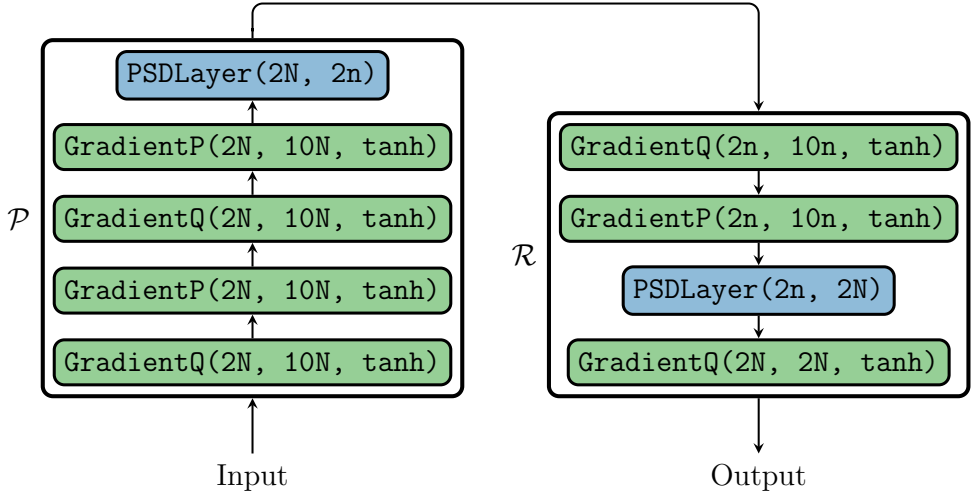

We train on data coming from 20 different choices of the parameter $\mu$ that are equally spaced in the interval $\mathbb{P} = [5/12, 2/3]$; the training is done for 100 epochs. The precise architecture is shown in figure \ref{fig:encoder_decoder}.

This system was already treated with so-called ``weakly-symplectic autoencoders'' (\cite{doi:10.1137/21M1466657}). These do not enforce symplecticity through the neural network architecture, as we do it here, but attempt to take symplecticity into account through an extra term in the loss function (PiNN approach, see \cite{raissi2019physics}). This has obvious pitfalls as there are no guarantees on symplecticity which lead to unpredictable behaviour in the online stage that is also discussed in \cite{doi:10.1137/21M1466657}.

As was done in \cite{doi:10.1137/21M1466657}, we compare two errors: the \textit{projection error} and the \textit{reduction error}; these are: 

\begin{equation}
e_\mathrm{proj}(\mu) := 
	\sqrt{
    \frac{\sum_{t=0}^K|| \mathbf{x}^{(t)}(\mu) - \mathcal{R}\circ\mathcal{P}(\mathbf{x}^{(t)}(\mu)) ||^2}{\sum_{t=0}^K|| \mathbf{x}^{(t)}(\mu) ||^2}
},
	\label{eq:projection_error}
\end{equation}

and 

\begin{equation}
e_\mathrm{red}(\mu) := \sqrt{
    \frac{\sum_{t=0}^K|| \mathbf{x}^{(t)}(\mu) - \mathcal{R}(\mathbf{x}^{(t)}_r(\mu)) ||^2}{\sum_{t=0}^K|| \mathbf{x}^{(t)}(\mu) ||^2}
},
\end{equation}

where $||\cdot||$ is the $L_2$ norm (one could also optimize for different norms), $\mathbf{x}^{(t)}$ is the solution of the FOM at point $t$ and $\mathbf{x}^{(t)}_r$ is the solution of the ROM (in the reduced basis) at point $t$. This is obtained by integraing the reduced system (second vector field in equation \eqref{eq:reduced_vector_field}).
The projection error computes how well a reduced basis, represented by the reduction $\mathcal{P}$ and the reconstruction $\mathcal{R}$, can represent the data with which it is built. The reduction error on the other hand measures how far the reduced system diverges from the full-order system during integration (online stage).

\begin{figure}
	\begin{subfigure}{.49\textwidth}
		\includegraphics[width=\textwidth]{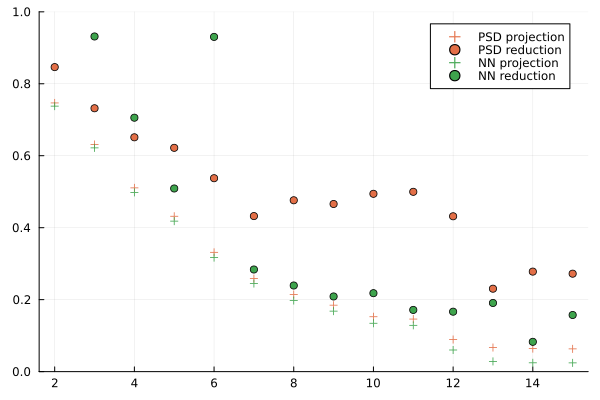}
		\caption{$\mu=0.47$}
	\end{subfigure}
	\begin{subfigure}{.49\textwidth}
		\includegraphics[width=\textwidth]{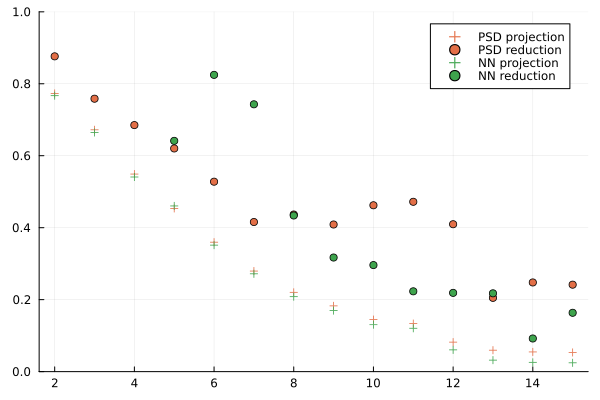}
		\caption{$\mu=0.51$}
	\end{subfigure}
	\begin{subfigure}{.49\textwidth}
		\includegraphics[width=\textwidth]{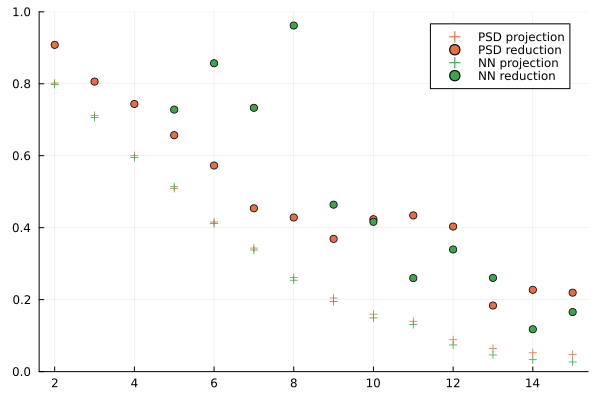}
		\caption{$\mu=0.55$}
	\end{subfigure}
	\begin{subfigure}{.49\textwidth}
		\includegraphics[width=\textwidth]{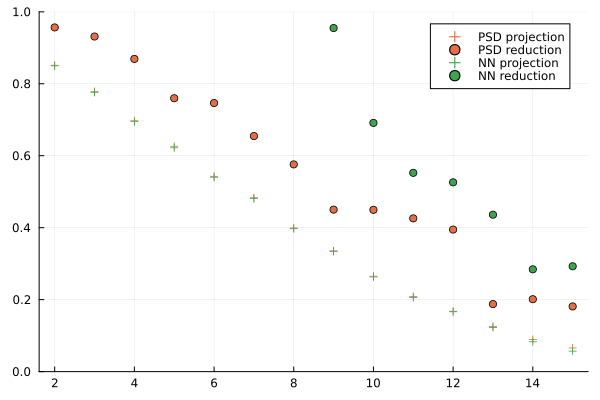}
		\caption{$\mu=0.625$}
	\end{subfigure}

	\caption{Projection and reduction errors for the different choices of the parameter $\mu$.}
	\label{fig:projection_reduction_errors}
\end{figure}

Training of the neural networks is done with \texttt{GeometricMachineLearning.jl}\footnote{\href{https://github.com/JuliaGNI/GeometricMachineLearning.jl}{https://github.com/JuliaGNI/GeometricMachineLearning.jl}}. This contains almost all relevent components for optimizing the neural network shown in figure \ref{fig:encoder_decoder}, including the generalization of common optimizers to manifolds. (It does however not include an \textit{automatic differentiation routine}. This is taken from the Julia package \texttt{Zygote.jl}\footnote{\href{https://github.com/FluxML/Zygote.jl}{https://github.com/FluxML/Zygote.jl}}.)
As optimizer we pick the Adam optimzier (see \cite{kingma2014adam}) with the values $\eta=0.001$, $\rho_1=0.9$, $\rho_2=0.99$ and $\delta=10^{-8}$. The optimization for the PSD-like layers is done with the generalized optimizer introduced in \cite{brantner2023generalizing}. The data and network parameters are all of type \texttt{Float64}.

The encoder and decoder are evaluated for four different choices of the parameter $\mu$: 0.470, 0.510, 0.550 and 0.625; none of these appear in the training set. The results are shown in figure \ref{fig:projection_reduction_errors} including a comparison to PSD.

The symplectic autoencoders almost consistently outperform the PSD for all of these parameters, exept for $\mu=0.625$. We also observe a monotonous decrease in the reduction error as we increase the dimension of the reduced space. This was not observed for the ``weakly-symplectic autoencoders'' in \cite{doi:10.1137/21M1466657}, where increasing the dimension of the reduced space did not necessarily imply an improvement in the reduction error.

\section{Summary}

We introduced a novel neural network architecture that, for the first time, can map between spaces of different dimension while simultaneously preserving symplecticity. In order to achieve this we rely on manifold optimization and a generalized Adam optimizer.

The advantages of the new neural network design was demonstrated by comparing it to proper symplectic decomposition (PSD). PSD was up to this point the only data-driven method that can map between spaces of different dimension while also preserving symplecticity. PSD is however, unlike the neural network introduced in this paper, strictly limited to linear mappings.

\section*{Acknowledgements}

We would like to thank Tobias Blickhan for valuable discussions and Tatjana Stykel for reading an early version of the manuscript and providing valuable remarks that greatly improved the exposition of this work.

\appendix 

\section{Riemannian gradients for the Stiefel and the Symplectic Stiefel manifold}
\label{riemannian_gradient}

Here we check that the definitions for the Riemannian gradients in equation \eqref{eq:stsp_gradient} are the correct ones. The Riemannian metrics for $St(n, N)$ and $St(2n, 2N)$ are: 

\begin{align}
	g_Y(V_1,V_2) & = \mathrm{Tr}(V_1^T\left(\mathbb{I}_{2N} - \frac{1}{2}YY^T\right)V_2) & \text{ and} \\
	\hat{g}_Y(W_1, W_2) & = \mathrm{Tr}(W_1^T\left(\mathbb{I}_{2N} - \frac{1}{2}\mathbb{J}_{2N}^TY(Y^TY)^{-1}Y^T\mathbb{J}_{2N}\right)W_2(Y^TY)^{-1}).
\end{align}

For the first of these equations see e.g. \cite{brantner2023generalizing} and for the second see e.g. \cite{bendokat2021real}. See definition \ref{def:riemannian_gradient} for the Riemannian gradient. 

Here $\langle\cdot,\cdot\rangle$ is the \textit{dual pairing} between the tangent space and its dual. For a matrix manifold $\mathcal{M}$ and a function $L:\mathcal{M}\to\mathbb{R}$ we have: $\langle{}dL,V\rangle = \mathrm{Tr}((\nabla_YL)^TV)$.

It can now easily be verified that equation \eqref{eq:stsp_gradient} shows the correct Riemannian gradients. For the first case: 
\begin{align}
	g_Y(\mathrm{grad}_YL, V)  & =  \mathrm{Tr}((\nabla_YL)^TV - (Y^T\nabla{}L)Y^TV - \frac{1}{2}(\nabla_YL)^TYY^TV + \frac{1}{2}(Y^T\nabla{}L)Y^TYY^TV) \\
	& =  \mathrm{Tr}((\nabla_YL)^TV - (Y^T\nabla{}L)Y^TV - \frac{1}{2}(\nabla_YL)^TYY^TV + \frac{1}{2}(Y^T\nabla{}L)Y^TV) \\ 
	& = \mathrm{Tr}((\nabla_YL)^TV - \frac{1}{2}(Y^T\nabla{}L)Y^TV - \frac{1}{2}(\nabla_YL)^TYY^TV) \\
	& = Tr((\nabla_YL)^TV).
\end{align}

Here we have used that $Y^TV = - V^TY$ and $\mathrm{Tr}(AB) = \mathrm{Tr}(BA)$ for arbitrary (compatible) matrices $A$ and $B$ as well as $\mathrm{Tr}(A) = \mathrm{Tr}(A^T)$ for $A$ a square matrix. To sum up: 
\begin{align}
 -\mathrm{Tr}((Y^T\nabla_yL)Y^TV) - \mathrm{Tr}((\nabla_YL)^TYY^TV) & = - \mathrm{Tr}((\nabla_YL)Y^TVY^T) + \mathrm{Tr}((\nabla_YL)^TYV^TY)  \\ 
 & = \mathrm{Tr}((\nabla_YL)Y^TVY^T) + \mathrm{Tr}(YV^TY(\nabla_YL)^T) = 0.
\end{align}
This concludes the proof that the first line in equation \eqref{eq:stsp_gradient} shows the correct Riemannian gradient for $St(n,N)$. With a similar calculation we can also do this for $Sp(2n, 2N)$ (also see \cite{bendokat2021real}).

\phantomsection
\addcontentsline{toc}{section}{References}
\bibliographystyle{plainnat}
\bibliography{symplectic_autoencoder.bib}

\begin{thebibliography}{34}
\providecommand{\natexlab}[1]{#1}
\providecommand{\url}[1]{\texttt{#1}}
\expandafter\ifx\csname urlstyle\endcsname\relax
  \providecommand{\doi}[1]{doi: #1}\else
  \providecommand{\doi}{doi: \begingroup \urlstyle{rm}\Url}\fi

\bibitem[Absil et~al.(2008)Absil, Mahony, and Sepulchre]{absil2008optimization}
P-A Absil, Robert Mahony, and Rodolphe Sepulchre.
\newblock \emph{Optimization algorithms on matrix manifolds}.
\newblock Princeton University Press, 2008.

\bibitem[Antoulas et~al.(2001)Antoulas, Sorensen, and
  Gugercin]{antoulas2000survey}
Athanasios~C Antoulas, Danny~C Sorensen, and Serkan Gugercin.
\newblock A survey of model reduction methods for large-scale systems.
\newblock \emph{Contemp.Math.}, 280:\penalty0 193--220, 2001.

\bibitem[Arnold(1978)]{arnold1978mathematical}
VI~Arnold.
\newblock Mathematical methods of classical mechanics.
\newblock \emph{Ann Arbor}, 1001:\penalty0 48109, 1978.

\bibitem[Bank et~al.(2020)Bank, Koenigstein, and Giryes]{bank2020autoencoders}
Dor Bank, Noam Koenigstein, and Raja Giryes.
\newblock Autoencoders.
\newblock \emph{arXiv preprint arXiv:2003.05991}, 2020.

\bibitem[Bendokat and Zimmermann(2021)]{bendokat2021real}
Thomas Bendokat and Ralf Zimmermann.
\newblock The real symplectic stiefel and grassmann manifolds: metrics,
  geodesics and applications.
\newblock \emph{arXiv preprint arXiv:2108.12447}, 2021.

\bibitem[Bishop and Goldberg(1980)]{bishop1980tensor}
Richard~L Bishop and Samuel~I Goldberg.
\newblock \emph{Tensor Analysis on Manifolds}.
\newblock Courier Corporation, 1980.

\bibitem[Blickhan(2023)]{blickhan2023registration}
Tobias Blickhan.
\newblock A registration method for reduced basis problems using linear optimal
  transport.
\newblock \emph{arXiv preprint arXiv:2304.14884}, 2023.

\bibitem[Brantner(2023)]{brantner2023generalizing}
Benedikt Brantner.
\newblock Generalizing adam to manifolds for efficiently training transformers.
\newblock \emph{arXiv preprint arXiv:2305.16901}, 2023.

\bibitem[Buchfink et~al.(2023)Buchfink, Glas, and
  Haasdonk]{doi:10.1137/21M1466657}
Patrick Buchfink, Silke Glas, and Bernard Haasdonk.
\newblock Symplectic model reduction of hamiltonian systems on nonlinear
  manifolds and approximation with weakly symplectic autoencoder.
\newblock \emph{SIAM Journal on Scientific Computing}, 45\penalty0
  (2):\penalty0 A289--A311, 2023.
\newblock \doi{10.1137/21M1466657}.
\newblock URL \url{https://doi.org/10.1137/21M1466657}.

\bibitem[Edelman et~al.(1998)Edelman, Arias, and Smith]{edelman1998geometry}
Alan Edelman, Tom{\'a}s~A Arias, and Steven~T Smith.
\newblock The geometry of algorithms with orthogonality constraints.
\newblock \emph{SIAM journal on Matrix Analysis and Applications}, 20\penalty0
  (2):\penalty0 303--353, 1998.

\bibitem[Fresca et~al.(2021)Fresca, Dede, and Manzoni]{fresca2021comprehensive}
Stefania Fresca, Luca Dede, and Andrea Manzoni.
\newblock A comprehensive deep learning-based approach to reduced order
  modeling of nonlinear time-dependent parametrized pdes.
\newblock \emph{Journal of Scientific Computing}, 87\penalty0 (2):\penalty0
  1--36, 2021.

\bibitem[Gao et~al.(2021)Gao, Son, Absil, and Stykel]{gao2021riemannian}
Bin Gao, Nguyen~Thanh Son, P-A Absil, and Tatjana Stykel.
\newblock Riemannian optimization on the symplectic stiefel manifold.
\newblock \emph{SIAM Journal on Optimization}, 31\penalty0 (2):\penalty0
  1546--1575, 2021.

\bibitem[Gao et~al.(2022)Gao, Son, and Stykel]{gao2022optimization}
Bin Gao, Nguyen~Thanh Son, and Tatjana Stykel.
\newblock Optimization on the symplectic stiefel manifold: Sr
  decomposition-based retraction and applications.
\newblock \emph{arXiv preprint arXiv:2211.09481}, 2022.

\bibitem[Goodfellow et~al.(2016)Goodfellow, Bengio, and
  Courville]{goodfellow2016deep}
Ian Goodfellow, Yoshua Bengio, and Aaron Courville.
\newblock \emph{Deep learning}.
\newblock MIT press, 2016.

\bibitem[Greif and Urban(2019)]{greif2019decay}
Constantin Greif and Karsten Urban.
\newblock Decay of the kolmogorov n-width for wave problems.
\newblock \emph{Applied Mathematics Letters}, 96:\penalty0 216--222, 2019.

\bibitem[Gromov(1985)]{gromov1985pseudo}
Mikhael Gromov.
\newblock Pseudo holomorphic curves in symplectic manifolds.
\newblock \emph{Inventiones mathematicae}, 82\penalty0 (2):\penalty0 307--347,
  1985.

\bibitem[Hairer et~al.(2006)Hairer, Lubich, and Wanner]{hairer2006geometric}
Ernst Hairer, Christian Lubich, and Gerhard Wanner.
\newblock \emph{Geometric Numerical integration: structure-preserving
  algorithms for ordinary differential equations}.
\newblock Springer, 2006.

\bibitem[Hornik et~al.(1989)Hornik, Stinchcombe, and
  White]{hornik1989multilayer}
Kurt Hornik, Maxwell Stinchcombe, and Halbert White.
\newblock Multilayer feedforward networks are universal approximators.
\newblock \emph{Neural networks}, 2\penalty0 (5):\penalty0 359--366, 1989.

\bibitem[Innes et~al.(2019)Innes, Edelman, Fischer, Rackauckas, Saba, Shah, and
  Tebbutt]{innes2019differentiable}
Mike Innes, Alan Edelman, Keno Fischer, Chris Rackauckas, Elliot Saba, Viral~B
  Shah, and Will Tebbutt.
\newblock A differentiable programming system to bridge machine learning and
  scientific computing.
\newblock \emph{arXiv preprint arXiv:1907.07587}, 2019.

\bibitem[Jin et~al.(2020)Jin, Zhang, Zhu, Tang, and
  Karniadakis]{jin2020sympnets}
Pengzhan Jin, Zhen Zhang, Aiqing Zhu, Yifa Tang, and George~Em Karniadakis.
\newblock Sympnets: Intrinsic structure-preserving symplectic networks for
  identifying hamiltonian systems.
\newblock \emph{Neural Networks}, 132:\penalty0 166--179, 2020.

\bibitem[Jin et~al.(2022)Jin, Lin, and Xiao]{jin2022optimal}
Pengzhan Jin, Zhangli Lin, and Bo~Xiao.
\newblock Optimal unit triangular factorization of symplectic matrices.
\newblock \emph{Linear Algebra and its Applications}, 2022.

\bibitem[Kingma and Ba(2014)]{kingma2014adam}
Diederik~P Kingma and Jimmy Ba.
\newblock Adam: A method for stochastic optimization.
\newblock \emph{arXiv preprint arXiv:1412.6980}, 2014.

\bibitem[Kong et~al.(2022)Kong, Wang, and Tao]{kong2022momentum}
Lingkai Kong, Yuqing Wang, and Molei Tao.
\newblock Momentum stiefel optimizer, with applications to suitably-orthogonal
  attention, and optimal transport.
\newblock \emph{arXiv preprint arXiv:2205.14173}, 2022.

\bibitem[Kraus(2013)]{kraus2013variational}
Michael Kraus.
\newblock Variational integrators in plasma physics.
\newblock \emph{arXiv preprint arXiv:1307.5665}, 2013.

\bibitem[Lassila et~al.(2014)Lassila, Manzoni, Quarteroni, and
  Rozza]{lassila2014model}
Toni Lassila, Andrea Manzoni, Alfio Quarteroni, and Gianluigi Rozza.
\newblock Model order reduction in fluid dynamics: challenges and perspectives.
\newblock \emph{Reduced Order Methods for modeling and computational
  reduction}, pages 235--273, 2014.

\bibitem[Lee and Carlberg(2020)]{lee2020model}
Kookjin Lee and Kevin~T Carlberg.
\newblock Model reduction of dynamical systems on nonlinear manifolds using
  deep convolutional autoencoders.
\newblock \emph{Journal of Computational Physics}, 404:\penalty0 108973, 2020.

\bibitem[Leimkuhler and Reich(2004)]{leimkuhler2004simulating}
Benedict Leimkuhler and Sebastian Reich.
\newblock \emph{Simulating hamiltonian dynamics}.
\newblock Cambridge university press, Cambridge, 2004.

\bibitem[McDuff and Salamon(2017)]{mcduff2017introduction}
Dusa McDuff and Dietmar Salamon.
\newblock \emph{Introduction to symplectic topology}, volume~27.
\newblock Oxford University Press, 2017.

\bibitem[Moses et~al.(2021)Moses, Churavy, Paehler, H{\"u}ckelheim, Narayanan,
  Schanen, and Doerfert]{moses2021reverse}
William~S Moses, Valentin Churavy, Ludger Paehler, Jan H{\"u}ckelheim, Sri
  Hari~Krishna Narayanan, Michel Schanen, and Johannes Doerfert.
\newblock Reverse-mode automatic differentiation and optimization of gpu
  kernels via enzyme.
\newblock In \emph{Proceedings of the international conference for high
  performance computing, networking, storage and analysis}, pages 1--16, 2021.

\bibitem[Peng and Mohseni(2016)]{peng2016symplectic}
Liqian Peng and Kamran Mohseni.
\newblock Symplectic model reduction of hamiltonian systems.
\newblock \emph{SIAM Journal on Scientific Computing}, 38\penalty0
  (1):\penalty0 A1--A27, 2016.

\bibitem[Raissi et~al.(2019)Raissi, Perdikaris, and
  Karniadakis]{raissi2019physics}
Maziar Raissi, Paris Perdikaris, and George~E Karniadakis.
\newblock Physics-informed neural networks: A deep learning framework for
  solving forward and inverse problems involving nonlinear partial differential
  equations.
\newblock \emph{Journal of Computational physics}, 378:\penalty0 686--707,
  2019.

\bibitem[Sanz-Serna and Calvo(2018)]{sanz2018numerical}
Jesus-Maria Sanz-Serna and Mari-Paz Calvo.
\newblock \emph{Numerical hamiltonian problems}, volume~7.
\newblock Dover Publications, Mineola, NY, 2018.

\bibitem[Tyranowski and Kraus(2022)]{tyranowski2019symplectic}
Tomasz~M Tyranowski and Michael Kraus.
\newblock Symplectic model reduction methods for the vlasov equation.
\newblock \emph{Contributions to Plasma Physics}, 2022.

\bibitem[Xu(2003)]{XU20031}
Hongguo Xu.
\newblock An svd-like matrix decomposition and its applications.
\newblock \emph{Linear Algebra and its Applications}, 368:\penalty0 1--24,
  2003.
\newblock ISSN 0024-3795.
\newblock \doi{https://doi.org/10.1016/S0024-3795(03)00370-7}.
\newblock URL
  \url{https://www.sciencedirect.com/science/article/pii/S0024379503003707}.

\end{thebibliography}

\end{document}